\documentclass[conference]{IEEEtran}
\IEEEoverridecommandlockouts
\usepackage{cite}
\usepackage{amsmath,amssymb,amsfonts}
\usepackage{algorithmic}
\usepackage{graphicx}
\usepackage{textcomp}
\usepackage{xcolor}
\usepackage{subfigure}
\usepackage{multirow}
\usepackage[linesnumbered,ruled,lined,commentsnumbered]{algorithm2e}
\graphicspath{{figure/}}
\def\BibTeX{{\rm B\kern-.05em{\sc i\kern-.025em b}\kern-.08em
    T\kern-.1667em\lower.7ex\hbox{E}\kern-.125emX}}
\begin{document}

\title{
	SRLF: A Stance-aware Reinforcement Learning Framework for Content-based Rumor Detection on Social Media
}

\author{
	Chunyuan Yuan\textsuperscript{\rm 1,2}, Wanhui Qian\textsuperscript{\rm 1,2}, Qianwen Ma\textsuperscript{\rm 1,2}, Wei Zhou\textsuperscript{\rm 1,*}, Songlin Hu\textsuperscript{\rm 1,2}\\
	\textsuperscript{\rm 1}Institute of Information Engineering, Chinese Academy of Sciences, Beijing, China\\
	\textsuperscript{\rm 2}School of Cyber Security, University of Chinese Academy of Sciences, Beijing, China\\  
	 \{yuanchunyuan,qianwanhui,maqianwen,zhouwei,husonglin\}@iie.ac.cn \\
}



\maketitle

\begin{abstract}
The\let\thefootnote\relax\footnotetext{* Corresponding author.} rapid development of social media changes the lifestyle of people and simultaneously provides an ideal place for publishing and disseminating rumors, which severely exacerbates social panic and triggers a crisis of social trust. Early content-based methods focused on finding clues from the text and user profiles for rumor detection. Recent studies combine the stances of users' comments with news content to capture the difference between true and false rumors. Although the user's stance is effective for rumor detection, the manual labeling process is time-consuming and labor-intensive, which limits the application of utilizing it to facilitate rumor detection.

In this paper, we first finetune a pre-trained BERT model on a small labeled dataset and leverage this model to annotate weak stance labels for users' comment data to overcome the problem mentioned above. Then, we propose a novel \textbf{S}tance-aware \textbf{R}einforcement \textbf{L}earning \textbf{F}ramework (SRLF) to select high-quality labeled stance data for model training and rumor detection. Both the stance selection and rumor detection tasks are optimized simultaneously to promote both tasks mutually. We conduct experiments on two commonly used real-world datasets. The experimental results demonstrate that our framework outperforms the state-of-the-art models significantly, which confirms the effectiveness of the proposed framework.
\end{abstract}


\section{Introduction}
The rapid development of social media platforms such as Facebook, Twitter, and Sina Weibo has facilitated the dissemination of information, but at the same time, it has also provided an ideal platform for publishing and disseminating rumors. According to the report of Pew Research Center, 57\% of people think the news they see on social media to be largely inaccurate.~\footnote{https://www.journalism.org/2018/09/10/news-use-across-social-media-platforms-2018/} The widespread rumors have seriously aggravated social panic and triggered a crisis of social trust. Therefore, it is critical to developing automatic methods to detect rumors.

\begin{figure}[htbp]
	\centering
	\includegraphics[scale=0.5]{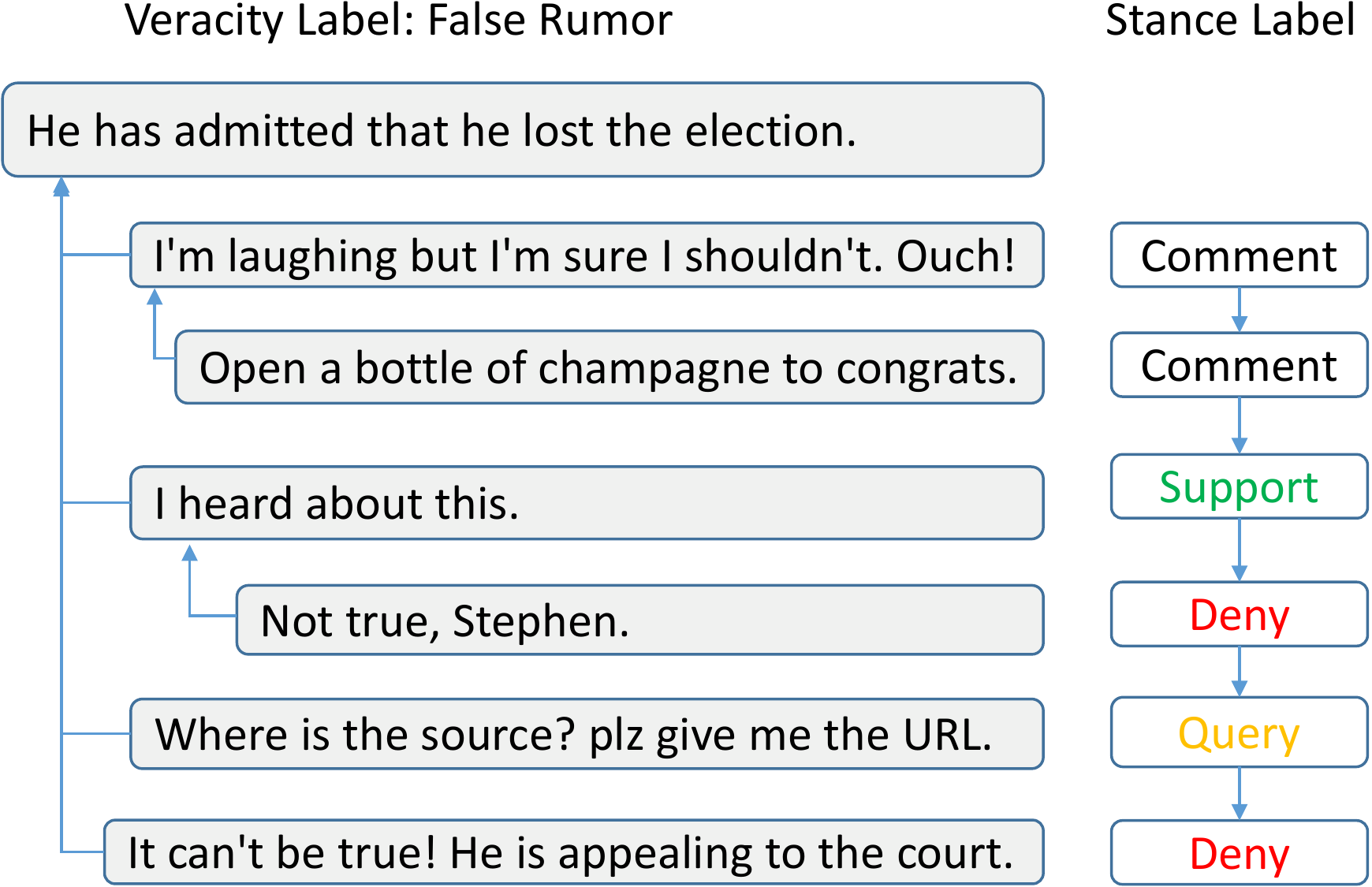}
	\caption{An example of twitter threads with stance and rumor veracity labels.}
	\label{example}
\end{figure}

Early studies mainly focused on designing effective features from various information sources, including text content~\cite{Castillo_2011,Qazvinian_2011,Popat_2017,yuan2019learning}, publisher's profiles~\cite{Castillo_2011, Yang_2012} and propagation patterns~\cite{jin2013epidemiological,sampson2016leveraging,ma2017detect}. However, manually designed features cannot capture the semantic representation of the news content and also easily manipulated by rumor posters, leading to poor detection performance.

Recent studies~\cite{rumor_yuan_2019,kumar2019tree,yu2020predicting,yuan2020early} can automatically learn semantic or structural representations from raw data and conduct deep feature interactions for rumor detection. For example, some studies~\cite{ma2018detect,kumar2019tree,yu2020predicting} combine the stance prediction and rumor detection tasks as a multi-task classification to improve the performance of rumor detection. As shown in Figure~\ref{example}, users' ``query'' and ``deny'' stances for the news provide effective signals to detect the veracity of the information. Although these methods achieve significant improvement on rumor detection, the stance labeling process of users' comments is time-consuming and labor-intensive. Furthermore, due to the dynamic nature of news, annotated users' stances of existing news documents are hard to be applied to emerged events, which requires continuous labeling of new events.

A possible solution is to train a model, such as BERT~\cite{devlin2019bert}, on a small labeled dataset for stance prediction and then use the model to automatically annotate the unlabeled comment data of users for every piece of news. However, weak stance labels produced by the pre-trained stance prediction model contain a lot of noises, which may affect the performance of rumor detection. To solve this limitation, we explore selecting high-quality stance labels from the weakly labeled data for facilitating rumor detection. We model the training process of the rumor detection and stance selection tasks as a reinforcement learning process to jointly optimize both tasks.

Reinforcement learning is usually used in sequential decision-making tasks, such as games and robot training, in which there is a determinate environment to give its state and provide a reward to an action. However, there is no such environment and agent in this task. Thus, the main challenges of this task include: (1) Who plays the environment and the agent role? (2) What are the state and action? (3) How the action influences the state of the environment? (4) What is an effective reward to drive the agent to optimize this task? 

In this paper, we propose a novel reinforcement learning framework to address these challenges. The framework consists of (1) text representation learning and rumor detection module, which is the environment; (2) the stance selection module, which is the agent. The environment treats the text (source tweet and comments) representation as its state and the probabilities of truth labels as a reward. The agent treats the retaining or removing the stance of the comments as the action. The agent gives actions to the environment. The environment changes its text representation learning module to produce a new state and calculate the probabilities of truth labels, and then return these signals to the agent. In this way, both the environment and agent will be trained together and finally improve both performances. We evaluate our proposed framework on two commonly used real-world datasets. The experimental results show that our method significantly outperforms several strong rumor detection baselines, which verifies the effectiveness of the proposed framework.

The contributions of this paper can be summarized as follows:
\begin{itemize}
	\item We put forward the time-consuming and label-intensive problem of manually labeling the user's stance for rumor detection, and provide a feasible framework SRLF for the problem.
	
	\item The framework can automatically annotate the unlabeled comment data and model the training process of the rumor detection and stance selection tasks as a reinforcement learning process, which can jointly optimize both tasks for improving both performances of rumor detection and stance selection tasks.
	
	\item We conduct extensive experiments on two real-world datasets. The results show that SRLF achieves superior improvements over state-of-the-art models on the rumor detection task. The source code will be released in the future for further research. 
\end{itemize}

The rest of the paper is organized as follows. In  Section~\ref{related_works_section}, we briefly review the related work. In Section~\ref{model_section}, we introduce the proposed model as well as its training strategy in detail. In Section~\ref{main_experiments}, we conduct experiments on two real-world datasets to evaluate the effectiveness of the proposed model on the rumor classification task. In Section~\ref{parameter_analysis}, we conduct a series of experiments to explore the influence of different hyper-parameters. Finally, we conclude with future work in Section~\ref{conclusion_section}.

\section{Related Work}  \label{related_works_section}
The target of this paper is to address the problems faced by content-based rumor detection methods. Therefore, we mainly introduce content-based methods proposed recently for rumor detection. These methods usually rely on relevant information (such as text content of news, users' comments, etc.) of news or tweets posted on social media platforms. Related research can be divided into the following categories: (1) Feature-based methods; (2) Deep learning methods.

\subsection{Feature-based Methods} 
Some early studies~\cite{Castillo_2011,Yang_2012,Kwon_2013,Ma_2015} focus on detecting rumors based on manually designed features. These features are mainly extracted from text content and users' profile data. Specifically, \cite{Castillo_2011} exploited various types of features, i.e., text-based, user-based, topic-based features, to study the credibility of news on Twitter. \cite{Kwon_2013} explored a novel approach to identify rumors based on temporal, structural, and linguistic properties of rumor propagation. \cite{Ma_2015} explored the temporal characteristics of these features based on the time series of rumor's life cycle to incorporate various social context information. 

However, the language used in social media is highly informal, which makes the designed features hardly capture the semantic information of the content of news. Furthermore, these features are easily manipulated by rumor posters when they found their previous tweets or accounts are blocked by the detection. These problems will significantly influence the performance of rumor detection.

\subsection{Deep Learning Methods}
To tackle the above problems of traditional feature-based methods, researchers apply deep learning-based models to learn efficient features automatically for rumor detection in recent years. Recurrent neural network (RNN)~\cite{ma2016detecting}, convolutional neural network (CNN)~\cite{Yu_2017} and graph neural network~\cite{rumor_yuan_2019} have been imported to learn the representations from news content or diffusion graph. Some studies also combine news content and users' response, such as conflicting viewpoints~\cite{jin2016news}, topics~\cite{guo2018rumor}, or stance~\cite{Bhatt_2018,li2019rumor}, to find clues by neural networks for fake news detection. To further improve the detection performance, some studies~\cite{ma2018detect,kumar2019tree,yu2020predicting} explored combining the stance prediction and rumor detection tasks as a multi-task classification. \cite{wang2020weak} proposed a reinforcement learning based framework to enlarge the amount of training data for fake news detection.

However, the manual labeling process of users' comments is time-consuming and labor-intensive. Furthermore, due to the dynamic nature of news, annotated users' stances of existing news documents are impossible to be applied to newly emerged events, which requires continuous labeling of new events. These problems will lead to the high cost of using users' stance for facilitating rumor detection.

\begin{figure*}[!htbp]
	\centering
	\includegraphics[scale=0.75]{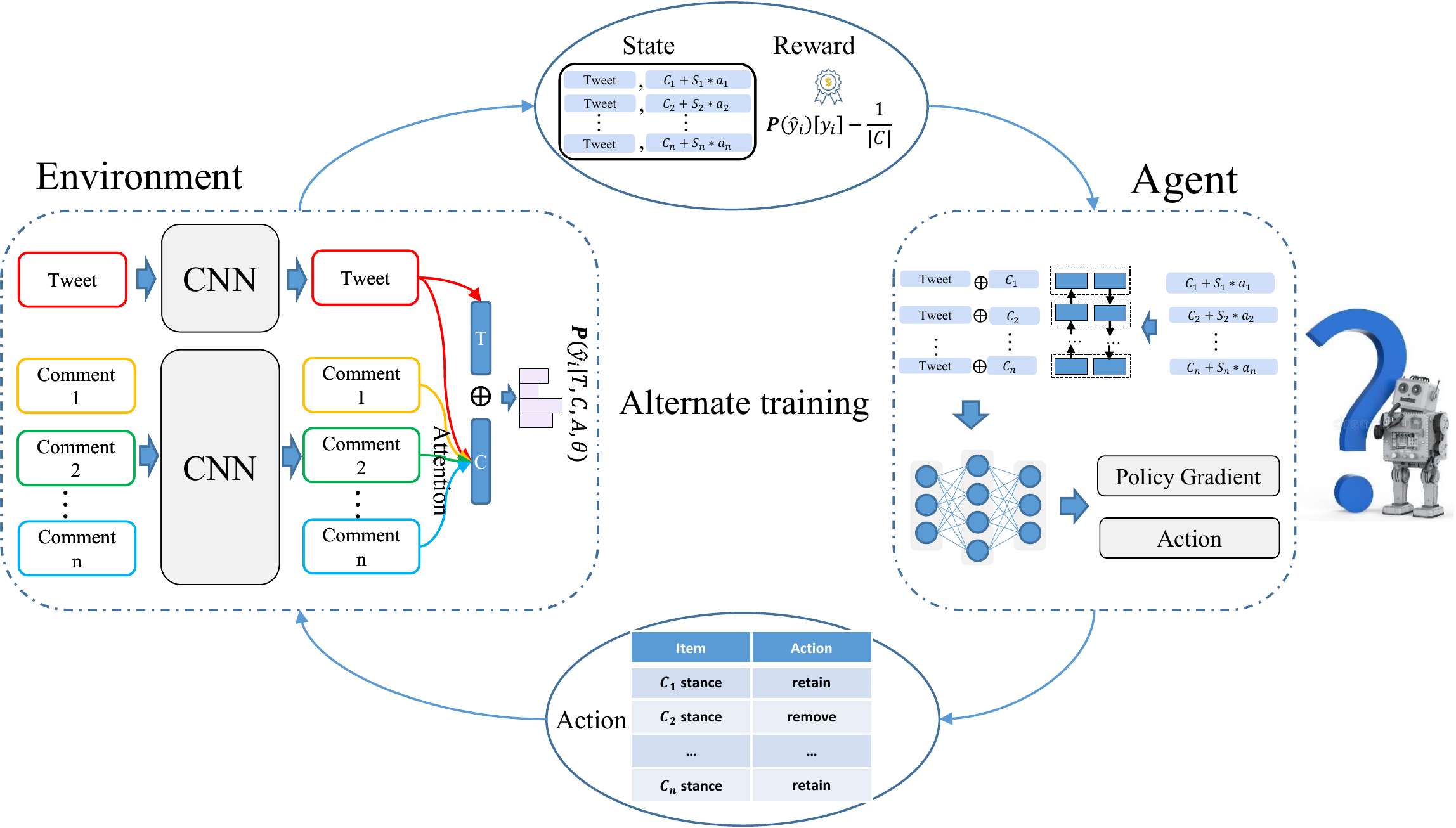}
	\caption{The structure of the stance-aware reinforcement learning framework for rumor detection.}
	\label{framework}
\end{figure*}

\section{Model} \label{model_section}
In this paper, we model the training process of both the rumor detection and stance selection tasks as a reinforcement learning process. The major components consist of Environment, which receives actions then provides reward and its new state; and agent, which receives rewards and state from the environment, then takes actions. In the following subsection, we will introduce: (1) Who plays the environment and the agent role? (2) What are the state and action? (3) How action influences the state of the environment? (4) What is an effective reward to drive the agent to optimize this task? Figure~\ref{framework} illustrates the architecture of the proposed framework.

\subsection{Annotation of the Comments}
Firstly, we utilize a small part of comment data with labeled stance to finetune a BERT model. Then, we apply the BERT model to annotate the rest unlabeled comments to produce the weak stance label for them. The details can be found in Section~\ref{dataset_section}.

\subsection{Environment}
The task of the environment is to receive the action from the agent and change its state and provide a reward to the agent. We introduce this process in detail.

\subsubsection{Text Representation}
We treat $\mathbf{x_j} \in \mathbb{R}^d$ as the $d$-dimensional embedding of the $j$-th word in the news content $\mathbf{t_i}$. We suppose every piece of news has $L$ words. We truncate the news at the end position if it is longer than $L$ and pad zero at the start of it when the length of the news is shorter than $L$.

A document of length $L$ can be represented as 
\begin{equation}
	\mathbf{x}_{1:L}= [\mathbf{x}_1 ; \mathbf{x}_2 ; \cdots ; \mathbf{x}_L] \,,  \\
\end{equation}
where `;' is the concatenation operator. We use $\mathbf{x}_{j:j+k}$ to represent the concatenation of words $\mathbf{x}_j, \mathbf{x}_{j+1}, \ldots, \mathbf{x}_{j+k}$.

Same as previous studies~\cite{rumor_yuan_2019,wang2020weak,yuan2020early}, we employ the convolutional neural network~\cite{kim2014convolutional} as the basic module of the framework to learn the semantic representation of news content. Specifically, given the word embeddings $\left( \mathbf{x}_1, \mathbf{x}_2, \ldots, \mathbf{x}_L \right) \in \mathbb{R}^{L \times d} $ of the source news, we apply several convolution layers on the word embedding matrix:
\begin{equation}
	e_i = \mathbf{\mathop{ReLU}} \left(\mathbf{W} * \mathbf{x}_{i:i+h-1} \right) \,,
\end{equation}
to extract feature map $\mathbf{e} = [e_1, e_2, \ldots, e_{L-h+1}] \in \mathbb{R}^{L-h+1}$, where $\mathbf{W} \in \mathbb{R}^{h \times d}$ is the convolutional kernel with $h$ size of receptive filed. Then, we apply a max-overtime pooling operation over the feature map: $ \hat{e} = \max(\mathbf{e})$ to extract the most important feature from it. 

The CNN layer uses $d/3$ filters (with different kernel size $h \in \{3,4,5\}$ ) to extract multiple features from news content. Then, we stack all kinds of kernels' outputs to produce $\mathbf{t}_i \in \mathbb{R}^{d}$ as the embedding of the $i$-th source news $t_i$. 

In the same way, we can get text representation of every comment $c_n, n \in [1, N]$ for the source news. The comments' representations are stack together to form the comment matrix $\mathbf{C} = [\mathbf{c}_1; \mathbf{c}_2; \ldots; \mathbf{c}_N] \in \mathbb{R}^{N \times d}$.

\subsubsection{State}
The state is the basis of rumor detection for environment and stance selection for the agent. The state vector mainly consists of two components: the representation of source tweets and replies. 

Supposing the environment has received the action from the agent, then the environment will play the action to change its state. The given action is defined as retaining or removing the stance of the comments. We design a way to meet this process:
\begin{equation}
	\mathbf{\widetilde{c}}_n = \mathbf{c}_n + \mathbf{s}_n \cdot \mathbf{A}_n  \,,  n \in [0, N] \\
\end{equation}
where $\mathbf{s}_n \in \mathbb{R}^{d}$ is the stance embedding of the comment $c_n$, which is randomly initialized by a normal distribution. The action $\mathbf{A}_n \in \{0, 1\}$, where `0' denotes removing the stance of the comment and `1' denotes retaining it. 

Finally, the state of the environment after receiving the action from the agent can be defined as $(\mathbf{t}_i, [\mathbf{\widetilde{c}}_1, \mathbf{\widetilde{c}}_2, \ldots, \mathbf{\widetilde{c}}_N])$. For the initial state of the environment, we directly set all actions $\mathbf{A}_n$ to `0', which means the initial state is $(\mathbf{t}_i, [\mathbf{c}_1, \mathbf{c}_2, \ldots, \mathbf{c}_N])$.

\subsubsection{Rumor Detector}
After obtaining the state representation, we can use it to detect whether the tweet is a false rumor. Firstly, we apply an attention module to aggregate $N$ comments' representation as a feature vector. Then, we fuse the tweet and comment representation for classification. 

Specifically, an attention module is applied on comment representation $\mathbf{C} = [\mathbf{\widetilde{c}}_1, \mathbf{\widetilde{c}}_2, \ldots, \mathbf{\widetilde{c}}_N]$ and tweet representation $\mathbf{t}_i$:
\begin{equation}
	\begin{split}
		& \mathbf{Z}_1 = \mathbf{ReLU}( [\mathbf{T}_i, \mathbf{C}, \mathbf{T}_i - \mathbf{C}, \mathbf{T}_i \odot \mathbf{C}] \mathbf{W}_1 + b_1 ) \,, \\
		& \mathbf{\alpha} = \mathbf{ReLU}( \mathbf{Z}_1 \mathbf{W}_2  + b_2 ) \,, \\ 
	\end{split}
\end{equation}
where $\odot$ denotes the element-wise multiplication. $\mathbf{T}_i = [\mathbf{t}_i, \mathbf{t}_i, \ldots, \mathbf{t}_i]$ is $N$ copies of $\mathbf{t}_i$. $\mathbf{W}_1, \mathbf{W}_2$ are trainable matrices and $b_1, b_2$ are bias terms. 

The final comment vector can be represented as:
\begin{equation}
	\mathbf{\widetilde{c}} = \sum_{n=1}^{N} \mathbf{\alpha}_n \mathbf{C}_n  \,.  \\
\end{equation}
Then, we concatenate the tweet representation with the comment vector for rumor detection:
\begin{equation}
	\begin{split}
		& \mathbf{Z}_2 = \mathbf{ReLU}( [\mathbf{t}_i, \mathbf{\widetilde{c}}] \mathbf{W}_3 + b_3 ) \,, \\
		& P(c_i| T_i, C, A, \mathcal{\theta}_1) = \mathbf{softmax}(\mathbf{Z}_2 \mathbf{W}_4 + b_4 ) \,, \\
	\end{split}
\end{equation}
where $\mathbf{W}_3, \mathbf{W}_4$ are trainable matrices. $\theta_1$ denotes all parameters used in the environment.

We apply the cross-entropy loss as the optimization objective:
\begin{equation}
	\begin{split}
		& \mathcal{L}_{rumor} = -\sum_{i} y_i \log P(c_i| T_i, C, A, \mathcal{\theta}_1) + \frac{\lambda}{2} ||\mathcal{\theta}_1||^2_2 \,, \\
	\end{split}
	\label{loss1}
\end{equation}
where $y_i$ is the true label of sample $i$. We apply $\ell_2$ regularization on all parameters of the model to overcome overfitting problem. $\lambda$ is a regularization factor.

\subsubsection{Reward Function}
The target of the action is to retain the stances of the comments that can bring improvement to rumor detection. Therefore, we use the probability of true label as the reward to drive the agent. Considering the probability is always larger than zero, which cannot provide negative signal, we finally define the reward as $R_k = P(c_i| T_i, C, A, \mathcal{\theta}_1) - \frac{1}{r}$ at episode step $k$, where $r$ is the number of rumor class.

\subsection{Agent}

\subsubsection{Policy Network and Action}
The objective of the policy network is to automatically determine whether to retain or remove the weak stance labels of the comments for every tweet. The input of the policy network is the state of the environment. 

As illustrated in Figure~\ref{example}, there are dependency relations among different comments. This kind of sequential relation can help to facilitate the decision-making of the action. To learn the dependency relations, we apply bidirectional LSTM on the comment sequence $\mathbf{C} = [\mathbf{\widetilde{c}}_1, \mathbf{\widetilde{c}}_2, \ldots, \mathbf{\widetilde{c}}_N]$, which can be formulated as follows:
\begin{equation}
	\begin{split}
		\mathbf{\widetilde{C}} = [\overrightarrow{\mathbf{LSTM}}(\mathbf{C}), \overleftarrow{\mathbf{LSTM}}(\mathbf{C})]  \,. \\
	\end{split}
\end{equation}

Subsequently, we concatenate the tweet representation with every comment representation as the final feature for decision making, which can be formulated as follows:
\begin{equation}
	\small
	\begin{split}
		& \mathbf{Z}_3 = \mathbf{ReLU}([\mathbf{T}_i, \mathbf{\widetilde{C}}] \mathbf{W}_5 + b_5 )  \,,  \\
		& P(A_k | T_i, C, \mathcal{\theta}_2) = \mathbf{softmax}(\mathbf{Z}_3 \mathbf{W}_6 + b_6 ) \,, \\
	\end{split}
\end{equation}
where $\mathbf{W}_5 \in \mathbb{R}^{2d \times d}$ and $\mathbf{W}_6 \in \mathbb{R}^{d \times 2}$ are trainable matrices. Then, the action $A_k$ is sampled according to $P(A_k | T_i, C_n, \mathcal{\theta}_2)$. $k$ denotes the $k$-th episode step.

\subsubsection{Objective Function}
The selection criteria are based on whether retaining the stance of the comment can improve the performance of the rumor detector and the reward reflects how good is the selection decision. Thus, we aim to maximize the expected total reward to optimize the selection strategy. The objective function can be defined as: 
\begin{equation}
	J(\mathcal{\theta}_2) = \sum_{n=1}^{N} \sum_{k=1}^{K} P(A^{(n)}_k | T_i, C_n, \mathcal{\theta}_2) \cdot \widetilde{R}^{(n)}_k  \,, 
\end{equation}
where $\mathcal{\theta}_2$ denotes all parameters used in the policy network. $\widetilde{R}^{(n)}_k$ is the discount reward of comment $n$ at episode $k$, which can be calculated as: 
\begin{equation}
	\widetilde{R}^{(n)}_k = R^{(n)}_k + \gamma R^{(n)}_k + \ldots + \gamma^{K-k-1} R^{(n)}_k  \,,  
\end{equation}
where $\gamma$ is the discount factor.

We use the Monte-Carlo policy gradient strategy~\cite{sutton2018reinforcement} to update the parameters of the policy network. For each episode step $k$, we update the parameters by stochastic gradient ascent as follows:
\begin{equation}
	\mathcal{\theta}_2 = \mathcal{\theta}_2 + \alpha  \sum_{n=1}^{N} \widetilde{R}^{(n)}_k \nabla_{\mathcal{\theta}_2} log P(A^{(n)}_k | T_i, C_n, \mathcal{\theta}_2)  \,, 
	\label{policy_loss}
\end{equation}
where $\alpha$ is the learning rate.

In this paper, because the environment part also needs to train, so we utilize an alternative training strategy to train the environment and agent by turns. The complete algorithm is shown in Algorithm~\ref{alg_global}.

\begin{algorithm}[!htb] 
	\caption{The training process of the framework.}
	\small
	\label{alg_global} 
	\KwIn{ \\
		\quad 1. The text data of the source tweet and its comments; \\
		\quad 2. The ground truth lable of source tweet and weak stance label of comments.  \\
	}
	
	\For{epoch $t \in [0, 1, \ldots, T]$}{
		\For{sample $i \in [0, 1, \ldots, I] $}
		{
			\If{i is even}{
				Train the environment and update parameter $\mathcal{\theta}_1$.
			} \Else{
				\For{episode $k \in [0, 1, \ldots, K]$}{
					(1) Perform the agent-environment interaction and collect the reward. \\
					(2) Update the policy network parameters $\mathcal{\theta}_2$ according to Eq.\ref{policy_loss}
				}
			}
		}
	}
\end{algorithm}

\section{Experiment}  \label{main_experiments}

In this section, we will introduce the experiments to evaluate the effectiveness of SRLF. Specifically, we aim to answer the following evaluation questions:
\begin{itemize}
	\item EQ1: Can SRLF improve rumor detection performance by modeling the rumor detection and stance selection process as a reinforcement learning process? 
	
	\item EQ2: How effective is the rumor detection module and stance selection module of the SRLF, respectively, in improving the detection performance of SRLF?
	
	\item EQ3: What effect do the parameter settings have on the performance of the proposed framework SRLF?
\end{itemize}

\subsection{Dataset}  \label{dataset_section}
We evaluate the proposed framework on two commonly used real-world data collections: Twitter15~\cite{ma2017detect} and Twitter16~\cite{ma2017detect}. Table~\ref{tab1} shows the statistics of the two datasets. Both Twitter15 and Twitter16 datasets contain four categories ($r = 4$ in reward function), i.e., ``false rumor'' (FR), ``non-rumor'' (NR), ``unverified'' (UR), and ``debunk rumor'' (TR).  Since the original datasets do not contain the comment data, we crawled all the data via Twitter API\footnote{https://dev.twitter.com/rest/public}.

\begin{table}[!htbp]
	\centering
	\caption{Dataset statistics.}
	\setlength{\tabcolsep}{4mm}{
		\begin{tabular}{c|c c c}
			\hline
			\textbf{Statistic}&\textbf{Twitter15} &\textbf{Twitter16} \\
			\hline
			\hline
			\# source tweets&  1490 &818 \\
			\hline
			\# non-rumors&  374 &205 \\
			\hline
			\# false rumors&  370 &205 \\
			\hline
			\# unverified rumors& 374 &203 \\
			\hline
			\# true rumors&  372 &205 \\
			\hline
			\# users&  276,663 &173,487 \\
			\hline
			\# posts&  331,612 &204,820 \\
			\hline
		\end{tabular}
	}
	\label{tab1}
\end{table}

We use a well-known dataset from SemEval-2019 Task 7-A~\cite{gorrell_2019_semeval} to fine-tune a pre-trained BERT model. Every tweet in the dataset has been annotated stance label by experts. The dataset contains 6702 tweets for training and 1872 for testing. The macro F1 of the finetuned BERT is 49.1\% on the test set. Finally, we leverage the finetuned BERT to annotate the Twitter15 and Twitter16 datasets for our model training.

\begin{table*}[!htbp]
	\begin{minipage}[t]{0.5\linewidth}
		\centering
		\caption{Experimental results on Twitter15 dataset.}
		\setlength{\tabcolsep}{3mm}{
			\begin{tabular}{|c|c|cccc|}
				\hline
				\multicolumn{6}{|c|}{\textit{Twitter15}} \\ \hline
				\multirow{2}{*}{Method} & \multirow{2}{*}{Acc.} & NR & FR & TR & UR \\ \cline{3-6}
				&  & $F_1$ & $F_1$ & $F_1$ & $F_1$ \\
				\hline
				DTR    & 0.409     & 0.501     & 0.311 & 0.364 & 0.473  \\
				\hline
				DTC    & 0.454     & 0.733     & 0.355 & 0.317 & 0.415  \\
				\hline
				RFC    & 0.565     & 0.810     & 0.422 & 0.401 & 0.543  \\
				\hline
				SVM-RBF    & 0.318     & 0.455     & 0.037 & 0.218 & 0.225  \\
				\hline
				SVM-TS    & 0.544     & 0.796     & 0.472 & 0.404 & 0.483  \\
				\hline
				PTK    & 0.750     & 0.804     & 0.698 & 0.765 & 0.733  \\
				\hline
				\hline
				GRU    & 0.646     & 0.792     & 0.574 & 0.608 & 0.592  \\
				\hline
				RvNN    & 0.723     & 0.682     & 0.758 & 0.821 & 0.654  \\
				\hline
				PPC    & 0.842     & 0.811     & 0.875 & 0.818 & 0.790  \\
				\hline
				MT-ES   & 0.848     & 0.800   & 0.865   & 0.885   & \textbf{0.854}  \\
				\hline
				LAN    & 0.854     & 0.847    & 0.883  & 0.875 & 0.816 \\
				\hline
				\hline
				SRLF    & \textbf{0.890}     & \textbf{0.890}     & \textbf{0.910} & \textbf{0.919} & 0.842  \\
				\hline
			\end{tabular}
			\label{exp_results_on_twitter15}
		}
	\end{minipage} 
	\begin{minipage}[t]{0.5\linewidth}  
		\centering
		\caption{
			Experimental results on Twitter16 dataset. 
		}
		\setlength{\tabcolsep}{3mm}{
			\begin{tabular}{|c|c|cccc|}
				\hline
				\multicolumn{6}{|c|}{\textit{Twitter16}} \\ \hline
				\multirow{2}{*}{Method} & \multirow{2}{*}{Acc.} & NR & FR & TR & UR \\ \cline{3-6}
				&  & $F_1$ & $F_1$ & $F_1$ & $F_1$ \\
				
				\hline
				DTR    & 0.414     & 0.394     & 0.273 & 0.630 & 0.344  \\
				\hline
				DTC    & 0.465     & 0.643 & 0.393& 0.419 & 0.403  \\
				\hline
				RFC    & 0.585     & 0.752 & 0.415 & 0.547 & 0.563  \\
				\hline
				SVM-RBF    & 0.321     & 0.423 & 0.085 & 0.419 & 0.037 \\
				\hline
				SVM-TS    & 0.574     & 0.755 & 0.420 & 0.571 & 0.526  \\
				\hline
				PTK    & 0.732     & 0.740 & 0.709 & 0.836 & 0.686  \\
				\hline
				\hline
				GRU    & 0.633     & 0.772 & 0.489 & 0.686 & 0.593  \\
				\hline
				RvNN    & 0.737     & 0.662     & 0.743 & 0.835 & 0.708  \\
				\hline
				PPC     & 0.863     & 0.820 & \textbf{0.898} & 0.843 & 0.837  \\
				\hline
				MT-ES   & 0.864    & 0.820     & 0.844   & \textbf{0.925}   & 0.871  \\
				\hline
				LAN     & 0.853     & 0.822  & 0.809 & 0.918 & 0.860 \\
				\hline
				\hline
				SRLF    & \textbf{0.886}     & \textbf{0.879}     & 0.854 & 0.909 & \textbf{0.899}  \\
				\hline
			\end{tabular}
			\label{exp_results_on_twitter16}
		}
	\end{minipage}
\end{table*}

\subsection{Comparison Models}  \label{baselines}
We compare our model with a series of content-based rumor detection models as follows:

(1) Feature-based methods:
\begin{itemize}
	\item \textbf{DTC}~\cite{Castillo_2011}: A decision tree-based model that utilizes a group of manually designed features extracted from users' posting and re-posting (``re-tweeting'') behavior, the text of the news, and citations to external sources.
	
	\item \textbf{SVM-RBF}~\cite{Yang_2012}: A SVM model with RBF kernel that utilizes a combination of news content features, account features, and location features.
	
	\item \textbf{SVM-TS}~\cite{Ma_2015}: A SVM-based model that utilizes social context features and time series of news propagation to model the variation of news characteristics.
	
	\item \textbf{DTR}~\cite{Zhao_2015}: A decision-tree framework that leverages enquiry patterns (the signal tweets) and clusters similar posts, and finally ranks the clusters based on statistical features for rumor detection.
	
	\item \textbf{RFC}~\cite{Kwon_2017}: A random forest classifier that utilizes user, linguistic and temporal characteristics for rumor classification.
	
	\item \textbf{PTK}~\cite{ma2017detect}: A SVM classifier with a tree kernel that detects rumors by learning temporal-structure patterns from trees of source news and users' comments.
\end{itemize}

(2) Deep learning methods:
\begin{itemize}
	\item \textbf{GRU}~\cite{ma2016detecting}: A RNN-based model that learns temporal-linguistic patterns from user comments to capture the variation of contextual dependency of relevant tweets over time.
	
	\item \textbf{RvNN}~\cite{ma2018rumor}: A bottom-up and top-down tree-structured model based on recursive neural networks to learn the recursive tree structure of news' comment data for rumor detection on Twitter.
	
	\item \textbf{PPC}~\cite{liu2018early}: A recurrent and convolutional model that detects rumors through capturing the global and local variations of user characteristics along the users' comment sequence. The model also can extract temporal features from the comment sequence.
	
	\item \textbf{MT-ES}~\cite{ma2018detect}: A multi-task architecture based on RNNs for capturing shared features from both the stance prediction and rumor detection subtasks. 
	
	\item \textbf{LAN}: A variant of GLAN~\cite{rumor_yuan_2019} model that encodes source tweet and comments by CNN and multi-head attention and jointly optimizes the stance prediction and rumor detection subtasks. 
\end{itemize}

It is worth noting that we do not compare with propagation-based methods because our work focuses on solving the problem content-based methods faced and did not use the social network of users or the diffusion graph of news.

\subsection{Data Preprocessing}
For a fair comparison, we use the train, validation, and test set that are split by~\cite{ma2016detecting,ma2018rumor,rumor_yuan_2019}, where 10\% samples as the validation dataset, and split the rest for training and test set with a ratio of 3:1.

For Twitter15 and Twitter16 datasets, the words are segmented by white space. The max length of source tweet and comments are set to 50. All word embeddings of the model are initialized with the 300-dimensional word vectors, which is released by~\cite{rumor_yuan_2019}. Words that are not present at the set of pre-trained word vectors are initialized from a uniform distribution~\cite{glorot2010understanding}. We keep the word vectors trainable in the training process, and they can be fine-tuned for each task.

\subsection{Model Configuration}
For a fair comparison, we adopt the same evaluation metrics used in the prior work~\cite{liu2018early,ma2018rumor,rumor_yuan_2019}. Thus,  this paper adopts the F1 score on every class and the overall accuracy for evaluation metrics. 

Our model is implemented using PyTorch~\cite{paszke2017automatic}. The weights and bias of the proposed framework are updated by Adam algorithm~\cite{kingma2014adam} and the hyper-parameters of Adam, $\beta_1$ and $\beta_2$ are 0.9 and 0.999 respectively. The learning rate is initialized as $1e^{-3}$ and gradually decreased during the process of training. We select the best parameter configuration based on performance on the development set and evaluate the configuration on the test set. 

The kernel size of the convolutional neural network is set to (3, 4, 5). There are 100 kernels for each kind of kernel size. The hidden size of bidirectional LSTM is set to 150. The batch size of the training set is set to 64. The episode $K$ is set to 10. The regularization parameter $\lambda$ is chosen from $\{1e^{-8}, 1e^{-7}, \ldots, 1e^{-2} \}$ and finally set to $1e^{-5}$. The discount factor $\gamma$ is chosen from $[0.1, 0.2, \ldots, 0.9, 0.99]$.


\subsection{Experiment Results and Analysis}
To answer EQ1, we compare our model with several content-based methods introduced in Section~\ref{baselines} for rumor detection. The experimental results are shown in the Table~\ref{exp_results_on_twitter15} and \ref{exp_results_on_twitter16}. For a fair comparison, the experimental results of baseline models are directly cited from previous studies~\cite{ma2018rumor,liu2018early}. 

\noindent Referring to the table, we can observe that:
\begin{itemize}
	\item The feature-based methods show poor performance for the rumor detection task, which indicates that hand-crafted features can not effectively encode semantic information of news content. Among these baselines, SVM-TS and RFC perform relatively better because they use additional temporal or user profile features. 
	
	\item For deep learning methods, RvNN and PPC outperform the feature-based methods, which indicates deep learning models can learn better semantic representations from text content. Furthermore, these neural network models can perform deep feature interactions among different types of features to capture the difference between rumor and non-rumors.
	
	\item MT-ES and LAN models show better performance than other deep learning models. We think the main reason is that both of them use the weak stance label as a supervised signal, which can provide more effective information because users' stances usually can reveal the veracity of rumors. However, the weak stances label annotated by BERT is not always reliable, which may influence the performance of these models.
	
	\item SRLF achieves significant improvement over the state-of-the-art models on both datasets. Compared with MT-ES and LAN models, SRLF can select high-quality labeled stance data for training, which can overcome the influence of wrong labeled data, thus showing better performance. The alternate training process can optimize both tasks simultaneously and promote them mutually.
\end{itemize}

In conclusion, the SRLF significantly outperforms feature-based methods and deep learning models. Both the accuracy and F1 score of rumor detection obtain significant improvement over other models. Specifically, our framework achieves 3.6\% and 2.2\% absolute improvement of the accuracy compared with the best performance of baselines on Twitter15 and Twitter16 datasets respectively, which demonstrates the effectiveness of the proposed framework.

\section{Further Analysis}

\subsection{Ablation Study}
To answer EQ2, we further perform some ablation studies over the different modules of the proposed framework. The experimental results are demonstrated in Table~\ref{ablation_study}. The ablation studies are conducted in the following orders: 
\begin{itemize}
	\item \textbf{w/o PL}: Removing the objective function of the agent.  
	\item \textbf{w/o DL}: Removing the rumor detector loss from the environment. 
\end{itemize}

\begin{table}[!htbp]
	\centering
	\caption{
		Ablation studies on Twitter15 and Twitter16 datasets.
	}
	\setlength{\tabcolsep}{2mm}{
		\begin{tabular}{|l|c|c|cccc|}
			\hline
			Dataset & Model    & Acc.     & NR    & FR & TR & UR \\
			\hline
			\multirow{3}{1cm}{Twitter15} & SRLF    & \textbf{0.890}     & \textbf{0.899}     & \textbf{0.913} & \textbf{0.912} & 0.830  \\
			& w/o PL  & 0.866     & 0.872    & 0.881  & 0.901   & 0.812\\
			& w/o DL  & 0.241     & 0.283    & 0.157 & 0.323 & 0.000 \\
			\hline
			
			\multirow{3}{1cm}{Twitter16} & SRLF    & \textbf{0.886}     & \textbf{0.879}     & 0.854 & 0.909 & \textbf{0.899}  \\
			&  w/o PL  & 0.859     & 0.835    & 0.800   & 0.918 & 0.876 \\
			&  w/o DL  & 0.272     & 0.000    & 0.169   & 0.404 & 0.080 \\
			\hline
		\end{tabular}
		\label{ablation_study}
	}
\end{table}

\noindent Referring to the experimental results of ablation studies in Table~\ref{ablation_study}, we can observe that:
\begin{itemize}
	\item We first examine the influence of the stance selection module (the agent). We can see that removing the objective function of the agent significantly affects performance, and the accuracy drops 2.4\% and 2.7\% on Twitter15 and Twitter16 datasets, respectively. The experimental results show that there are many noises in the weak stance labels of comments and removing the objective function makes the Agent unable to select high-quality stance labels for rumor detection.
	
	\item Then, we evaluate the influence of the rumor detector loss function on the performance. Referring to Table~\ref{ablation_study}, the detection performance decays to a very low level after removing the rumor detector loss. The rumor detector loss is applied to optimize the parameters of the environment. Removing it will cause the environment cannot to be trained, thus leading to very poor rumor detection performance.
\end{itemize}

In conclusion, both the rumor detection and stance selection subtasks are important for improving the performance of rumor detection. Thus, it is necessary to model the training process of both the rumor detection and stance selection tasks as a reinforcement learning process to jointly train both tasks and optimize them together.

\subsection{Parameter Sensitivity}  \label{parameter_analysis}
\begin{figure}[!htbp]
	\centering 
	\subfigure{
		\label{discount_factor} 
		\includegraphics[scale=0.5]{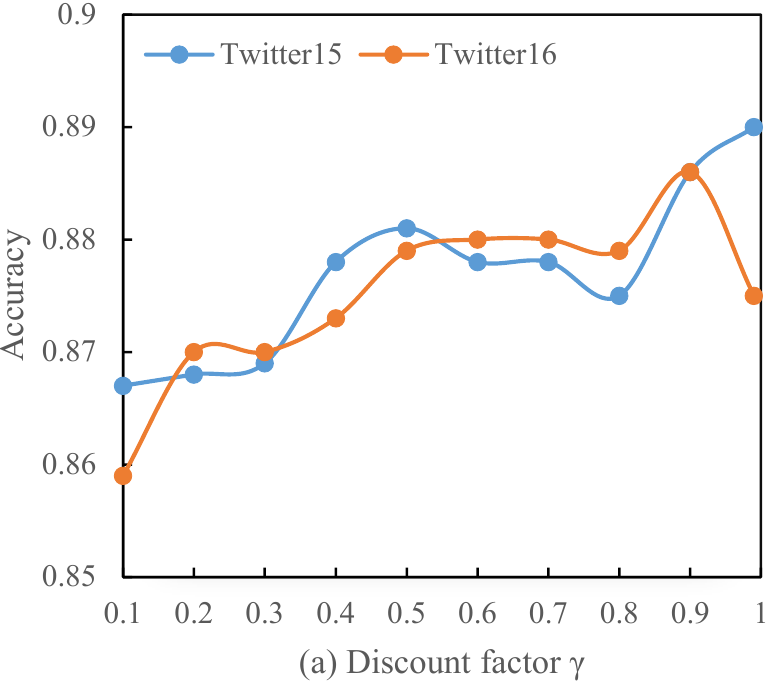} 
	}
	\subfigure{
		\label{regularization_factor} 
		\includegraphics[scale=0.5]{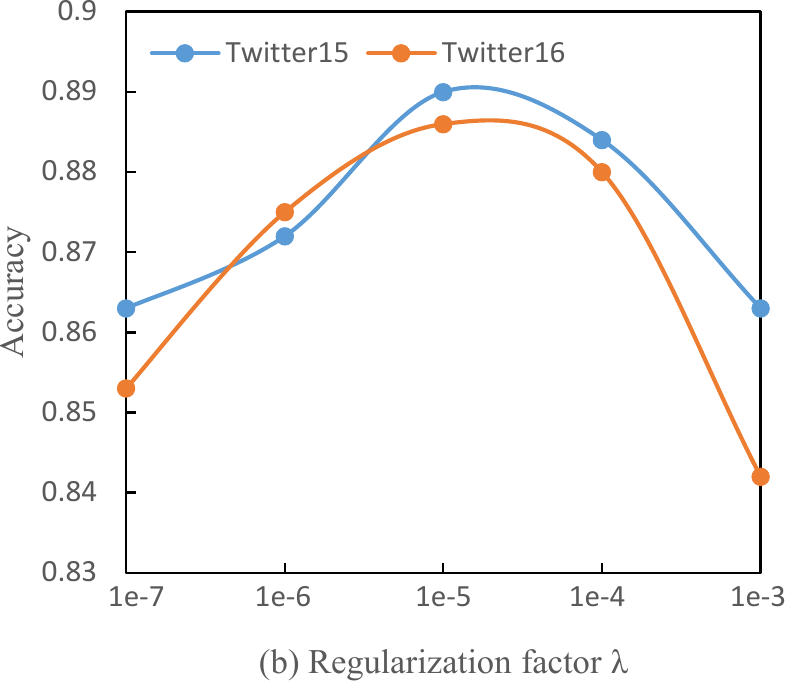} 
	}
	\caption{Effect of different (a) discount factors $\gamma$ and (b) regularization factors $\lambda$ on the datasets. }
	\label{parameter_sens}
\end{figure}

To answer EQ3, we conduct a series of parameter sensitivity experiments on two critical hyper-parameters with different settings. The experimental results are shown in Figure~\ref{parameter_sens}. 


\textbf{Impact of different discount factors $\gamma$:} Figure~\ref{discount_factor} shows the performance of the proposed framework with different discount factors $\gamma$. The discount factor relates the rewards to the time domain. A smaller $\gamma$ will lead the model to care less about the future reward, which may cause the agent unable to get proper training. Referring to the figure, the performance decays significantly when $\gamma$ decreases. The most suitable $\gamma$ is in the range from 0.9 to 0.99.

\textbf{Impact of different regularization factors $\lambda$: } Figure~\ref{regularization_factor} shows the performance with different regularization factors $\lambda$. The hyper-parameter $\lambda$ is applied to reduce the overfitting of the SRLF. An increase of $\lambda$ will result in a reduction of overfitting, but it will also bring a greater deviation. When $\lambda$ is too big or too small, the capacity of SRLF is severely restricted or easy to overfitting to the training data, leading to very poor performance finally. There is a balance between variance and bias, in which the model can achieve the best performance. Referring to the figure, the most suitable $\lambda$ is about $1e^{-5}$.

\section{Conclusion and Future Work} \label{conclusion_section}
Recent content-based rumor detection methods combine the users' stances data with news text content for rumor detection to improve the performance of rumor detection. However, the stance labeling process is very time-consuming and labor-intensive, which limits the application of using the stance of users to facilitating rumor detection. This paper proposes a novel stance-aware reinforcement learning framework to automatically annotate the weak stance of users' comments and select a high-quality stance for training rumor classifiers. In this way, we can address the above problem effectively. Extensive experiments are conducted on two commonly used real-world datasets. The experimental results show that our framework can outperform other state-of-the-art models significantly and thus confirms the effectiveness of the proposed framework. 

Since our model only applies the text content data of news and users' comment, it is complementary research compared with recent propagation-based methods. Therefore, it is promising to combine both ways to further improve the rumor detection performance.  For future work, we plan to integrate other types of information such as user profiles or diffusion structure of the news and social network of users into the framework to facilitate rumor detection.

\section*{Acknowledgment}
We gratefully thank the anonymous reviewers for their insightful comments. This research is supported in part by the National Key Research and Development Program of China under Grant 2018YFC0806900.


\clearpage

\bibliography{ijcnn.bib}

\begin{thebibliography}{10}
\providecommand{\url}[1]{#1}
\csname url@samestyle\endcsname
\providecommand{\newblock}{\relax}
\providecommand{\bibinfo}[2]{#2}
\providecommand{\BIBentrySTDinterwordspacing}{\spaceskip=0pt\relax}
\providecommand{\BIBentryALTinterwordstretchfactor}{4}
\providecommand{\BIBentryALTinterwordspacing}{\spaceskip=\fontdimen2\font plus
\BIBentryALTinterwordstretchfactor\fontdimen3\font minus
  \fontdimen4\font\relax}
\providecommand{\BIBforeignlanguage}[2]{{%
\expandafter\ifx\csname l@#1\endcsname\relax
\typeout{** WARNING: IEEEtran.bst: No hyphenation pattern has been}%
\typeout{** loaded for the language `#1'. Using the pattern for}%
\typeout{** the default language instead.}%
\else
\language=\csname l@#1\endcsname
\fi
#2}}
\providecommand{\BIBdecl}{\relax}
\BIBdecl

\bibitem{Castillo_2011}
C.~Castillo, M.~Mendoza, and B.~Poblete, ``Information credibility on
  twitter,'' in \emph{Proceedings of the 20th international conference on World
  wide web}.\hskip 1em plus 0.5em minus 0.4em\relax ACM, 2011, pp. 675--684.

\bibitem{Qazvinian_2011}
\BIBentryALTinterwordspacing
V.~Qazvinian, E.~Rosengren, D.~R. Radev, and Q.~Mei, ``Rumor has it:
  Identifying misinformation in microblogs,'' in \emph{Proceedings of the
  Conference on Empirical Methods in Natural Language Processing}, ser. EMNLP
  '11.\hskip 1em plus 0.5em minus 0.4em\relax Stroudsburg, PA, USA: Association
  for Computational Linguistics, 2011, pp. 1589--1599. [Online]. Available:
  \url{http://dl.acm.org/citation.cfm?id=2145432.2145602}
\BIBentrySTDinterwordspacing

\bibitem{Popat_2017}
\BIBentryALTinterwordspacing
K.~Popat, ``Assessing the credibility of claims on the web,'' in
  \emph{Proceedings of the 26th International Conference on World Wide Web
  Companion}, ser. WWW '17 Companion.\hskip 1em plus 0.5em minus 0.4em\relax
  Republic and Canton of Geneva, Switzerland: International World Wide Web
  Conferences Steering Committee, 2017, pp. 735--739. [Online]. Available:
  \url{https://doi.org/10.1145/3041021.3053379}
\BIBentrySTDinterwordspacing

\bibitem{yuan2019learning}
C.~Yuan, W.~Zhou, Q.~Ma, S.~Lv, J.~Han, and S.~Hu, ``Learning review
  representations from user and product level information for spam detection,''
  in \emph{2019 IEEE International Conference on Data Mining (ICDM)}.\hskip 1em
  plus 0.5em minus 0.4em\relax IEEE, 2019, pp. 1444--1449.

\bibitem{Yang_2012}
\BIBentryALTinterwordspacing
F.~Yang, Y.~Liu, X.~Yu, and M.~Yang, ``Automatic detection of rumor on sina
  weibo,'' \emph{Proceedings of the ACM SIGKDD Workshop on Mining Data
  Semantics - MDS ’12}, 2012. [Online]. Available:
  \url{http://dx.doi.org/10.1145/2350190.2350203}
\BIBentrySTDinterwordspacing

\bibitem{jin2013epidemiological}
F.~Jin, E.~Dougherty, P.~Saraf, Y.~Cao, and N.~Ramakrishnan, ``Epidemiological
  modeling of news and rumors on twitter,'' in \emph{Proceedings of the 7th
  Workshop on Social Network Mining and Analysis}.\hskip 1em plus 0.5em minus
  0.4em\relax ACM, 2013, p.~8.

\bibitem{sampson2016leveraging}
J.~Sampson, F.~Morstatter, L.~Wu, and H.~Liu, ``Leveraging the implicit
  structure within social media for emergent rumor detection,'' in
  \emph{Proceedings of the 25th ACM International on Conference on Information
  and Knowledge Management}.\hskip 1em plus 0.5em minus 0.4em\relax ACM, 2016,
  pp. 2377--2382.

\bibitem{ma2017detect}
J.~Ma, W.~Gao, and K.-F. Wong, ``Detect rumors in microblog posts using
  propagation structure via kernel learning,'' in \emph{Proceedings of the 55th
  Annual Meeting of the Association for Computational Linguistics (Volume 1:
  Long Papers)}, 2017, pp. 708--717.

\bibitem{rumor_yuan_2019}
C.~Yuan, Q.~Ma, W.~Zhou, J.~Han, and S.~Hu, ``Jointly embedding the local and
  global relations of heterogeneous graph for rumor detection,'' in \emph{The
  19th IEEE International Conference on Data Mining}.\hskip 1em plus 0.5em
  minus 0.4em\relax IEEE, 2019.

\bibitem{kumar2019tree}
S.~Kumar and K.~M. Carley, ``Tree lstms with convolution units to predict
  stance and rumor veracity in social media conversations,'' in
  \emph{Proceedings of the 57th Annual Meeting of the Association for
  Computational Linguistics}, 2019, pp. 5047--5058.

\bibitem{yu2020predicting}
J.~Yu, J.~Jiang, L.~M.~S. Khoo, H.~L. Chieu, and R.~Xia, ``Predicting stance
  and rumor veracity via dual hierarchical transformer with pretrained
  encoders,'' in \emph{Proceedings of the 2020 Conference on Empirical Methods
  in Natural Language Processing (EMNLP)}, 2020, pp. 1392--1401.

\bibitem{yuan2020early}
C.~Yuan, Q.~Ma, W.~Zhou, J.~Han, and S.~Hu, ``Early detection of fake news by
  utilizing the credibility of news, publishers, and users based on weakly
  supervised learning,'' in \emph{Proceedings of the 28th International
  Conference on Computational Linguistics}, 2020, pp. 5444--5454.

\bibitem{ma2018detect}
J.~Ma, W.~Gao, and K.-F. Wong, ``Detect rumor and stance jointly by neural
  multi-task learning,'' in \emph{Companion Proceedings of the The Web
  Conference 2018}, 2018, pp. 585--593.

\bibitem{devlin2019bert}
J.~Devlin, M.-W. Chang, K.~Lee, and K.~Toutanova, ``Bert: Pre-training of deep
  bidirectional transformers for language understanding,'' in \emph{NAACL-HLT
  (1)}, 2019.

\bibitem{Kwon_2013}
\BIBentryALTinterwordspacing
S.~Kwon, M.~Cha, K.~Jung, W.~Chen, and Y.~Wang, ``Prominent features of rumor
  propagation in online social media,'' \emph{2013 IEEE 13th International
  Conference on Data Mining}, Dec 2013. [Online]. Available:
  \url{http://dx.doi.org/10.1109/ICDM.2013.61}
\BIBentrySTDinterwordspacing

\bibitem{Ma_2015}
\BIBentryALTinterwordspacing
J.~Ma, W.~Gao, Z.~Wei, Y.~Lu, and K.-F. Wong, ``Detect rumors using time series
  of social context information on microblogging websites,'' \emph{Proceedings
  of the 24th ACM International on Conference on Information and Knowledge
  Management - CIKM ’15}, 2015. [Online]. Available:
  \url{http://dx.doi.org/10.1145/2806416.2806607}
\BIBentrySTDinterwordspacing

\bibitem{ma2016detecting}
J.~Ma, W.~Gao, P.~Mitra, S.~Kwon, B.~J. Jansen, K.-F. Wong, and M.~Cha,
  ``Detecting rumors from microblogs with recurrent neural networks,'' in
  \emph{Ijcai}, 2016, pp. 3818--3824.

\bibitem{Yu_2017}
\BIBentryALTinterwordspacing
F.~Yu, Q.~Liu, S.~Wu, L.~Wang, and T.~Tan, ``A convolutional approach for
  misinformation identification,'' \emph{Proceedings of the Twenty-Sixth
  International Joint Conference on Artificial Intelligence}, Aug 2017.
  [Online]. Available: \url{http://dx.doi.org/10.24963/ijcai.2017/545}
\BIBentrySTDinterwordspacing

\bibitem{jin2016news}
Z.~Jin, J.~Cao, Y.~Zhang, and J.~Luo, ``News verification by exploiting
  conflicting social viewpoints in microblogs,'' in \emph{Thirtieth AAAI
  conference on artificial intelligence}, 2016.

\bibitem{guo2018rumor}
H.~Guo, J.~Cao, Y.~Zhang, J.~Guo, and J.~Li, ``Rumor detection with
  hierarchical social attention network,'' in \emph{Proceedings of the 27th ACM
  International Conference on Information and Knowledge Management}, 2018, pp.
  943--951.

\bibitem{Bhatt_2018}
\BIBentryALTinterwordspacing
G.~Bhatt, A.~Sharma, S.~Sharma, A.~Nagpal, B.~Raman, and A.~Mittal, ``Combining
  neural, statistical and external features for fake news stance
  identification,'' \emph{Companion of the The Web Conference 2018 on The Web
  Conference 2018 - WWW ’18}, 2018. [Online]. Available:
  \url{http://dx.doi.org/10.1145/3184558.3191577}
\BIBentrySTDinterwordspacing

\bibitem{li2019rumor}
Q.~Li, Q.~Zhang, and L.~Si, ``Rumor detection by exploiting user credibility
  information, attention and multi-task learning,'' in \emph{Proceedings of the
  57th Annual Meeting of the Association for Computational Linguistics}, 2019,
  pp. 1173--1179.

\bibitem{wang2020weak}
Y.~Wang, W.~Yang, F.~Ma, J.~Xu, B.~Zhong, Q.~Deng, and J.~Gao, ``Weak
  supervision for fake news detection via reinforcement learning,'' in
  \emph{Proceedings of the AAAI Conference on Artificial Intelligence},
  vol.~34, no.~01, 2020, pp. 516--523.

\bibitem{kim2014convolutional}
Y.~Kim, ``Convolutional neural networks for sentence classification,'' in
  \emph{Proceedings of the 2014 Conference on Empirical Methods in Natural
  Language Processing (EMNLP)}, 2014, pp. 1746--1751.

\bibitem{sutton2018reinforcement}
R.~S. Sutton and A.~G. Barto, \emph{Reinforcement learning: An
  introduction}.\hskip 1em plus 0.5em minus 0.4em\relax MIT press, 2018.

\bibitem{gorrell_2019_semeval}
G.~Gorrell, E.~Kochkina, M.~Liakata, A.~Aker, A.~Zubiaga, K.~Bontcheva, and
  L.~Derczynski, ``{S}em{E}val-2019 task 7: {R}umour{E}val, determining rumour
  veracity and support for rumours,'' in \emph{Proceedings of the 13th
  International Workshop on Semantic Evaluation}.\hskip 1em plus 0.5em minus
  0.4em\relax Association for Computational Linguistics, Jun. 2019, pp.
  845--854.

\bibitem{Zhao_2015}
Z.~Zhao, P.~Resnick, and Q.~Mei, ``Enquiring minds: Early detection of rumors
  in social media from enquiry posts,'' in \emph{Proceedings of the 24th
  International Conference on World Wide Web}, ser. WWW'15.\hskip 1em plus
  0.5em minus 0.4em\relax Republic and Canton of Geneva, Switzerland:
  International World Wide Web Conferences Steering Committee, 2015, pp.
  1395--1405.

\bibitem{Kwon_2017}
\BIBentryALTinterwordspacing
S.~Kwon, M.~Cha, and K.~Jung, ``Rumor detection over varying time windows,''
  \emph{PLOS ONE}, vol.~12, no.~1, pp. 1--19, 01 2017. [Online]. Available:
  \url{https://doi.org/10.1371/journal.pone.0168344}
\BIBentrySTDinterwordspacing

\bibitem{ma2018rumor}
J.~Ma, W.~Gao, and K.-F. Wong, ``Rumor detection on twitter with
  tree-structured recursive neural networks,'' in \emph{Proceedings of the 56th
  Annual Meeting of the Association for Computational Linguistics (Volume 1:
  Long Papers)}, 2018, pp. 1980--1989.

\bibitem{liu2018early}
Y.~Liu and Y.-F.~B. Wu, ``Early detection of fake news on social media through
  propagation path classification with recurrent and convolutional networks,''
  in \emph{Thirty-Second AAAI Conference on Artificial Intelligence}, 2018.

\bibitem{glorot2010understanding}
X.~Glorot and Y.~Bengio, ``Understanding the difficulty of training deep
  feedforward neural networks,'' in \emph{Proceedings of the thirteenth
  international conference on artificial intelligence and statistics}.\hskip
  1em plus 0.5em minus 0.4em\relax JMLR Workshop and Conference Proceedings,
  2010, pp. 249--256.

\bibitem{paszke2017automatic}
A.~Paszke, S.~Gross, S.~Chintala, G.~Chanan, E.~Yang, Z.~DeVito, Z.~Lin,
  A.~Desmaison, L.~Antiga, and A.~Lerer, ``Automatic differentiation in
  pytorch,'' in \emph{NIPS-W}, 2017.

\bibitem{kingma2014adam}
D.~P. Kingma and J.~Ba, ``Adam: A method for stochastic optimization,''
  \emph{arXiv preprint arXiv:1412.6980}, 2014.

\end{thebibliography}
\bibliographystyle{IEEEtran}

\end{document}